\documentclass{article}

\usepackage{arxiv}
\usepackage{nicefrac}
\usepackage{microtype}
\usepackage{booktabs}
\usepackage[numbers,sort&compress]{natbib}
\usepackage{doi}
\usepackage{float}
\usepackage[T1]{fontenc}
\usepackage{graphicx}
\usepackage[utf8]{inputenc}

\title{Sales Research Agent \& Sales Research Bench}

\author{
  Deepanjan Bhol\\
  Senior Principal Engineering Lead\\
  Microsoft\\
  \texttt{debhol@microsoft.com}
}

\date{October 21, 2025}

\renewcommand{\headeright}{Research Article}
\renewcommand{\undertitle}{Enterprise AI for Sales}
\renewcommand{\shorttitle}{Sales Research Agent \& Sales Research Bench}

\hypersetup{
  pdftitle={Sales Research Agent \& Sales Research Bench Dynamics 365 IT Pro blog on 10/21},
  pdfsubject={q-bio.NC, q-bio.QM},
  pdfauthor={Deepanjan Bhol},
  pdfkeywords={},
}

\begin{document}
\maketitle

\begin{abstract}
The Sales Research Agent in Dynamics 365 Sales automatically connects to live CRM data and to additional data such as budgets and targets. It reasons over complex, customized schemas with deep domain expertise and produces decision-ready insights through narratives and rich visualizations tailored to the business question. But the market is crowded with AI offers whose quality is hard to verify, so Microsoft created the Sales Research Bench, a purpose-built benchmark that scores agents on eight customer-weighted dimensions (text and chart groundedness, relevance, explainability, schema accuracy, and chart quality). In a 200-question run on a customized enterprise schema (Oct 19, 2025), the Sales Research Agent outperformed Claude Sonnet 4.5 by 13 points and ChatGPT-5 by 24.1 points on the 100-point composite score, demonstrating enterprise readiness and giving customers a repeatable way to compare AI solutions. 
\end{abstract}

\section{Introduction}

The \textbf{Sales Research Agent} in Dynamics 365 Sales automatically connects to live CRM data and can connect to additional data stored elsewhere, such as budgets and targets. It reasons over complex, customized schemas, with deep domain expertise\textit{,} and presents novel, decision-ready insights through text-based narratives and rich data visualizations tailored to the business question at hand. 

For sales leaders, this means the ability to self-serve building out rich research journeys, spanning CRM and other domains, that previously took many people days or weeks to compile, with access to deeper insights enabled by the power of AI on pipeline, revenue attainment, and other critical topics.

But the market is crowded with offers that may or may not deliver acceptable levels of quality to support business decisions. How can business leaders know what’s truly enterprise ready? To help make sure customers do not have to rely on anecdotal evidence or ``gut feel", any vendor providing AI solutions must earn trust through clear, repeatable metrics that demonstrate quality, showing where the AI excels, where it needs improvement, and how it stacks up against alternatives.

\begin{figure}[H]
\centering
\includegraphics[width=13.76cm,height=6.84cm]{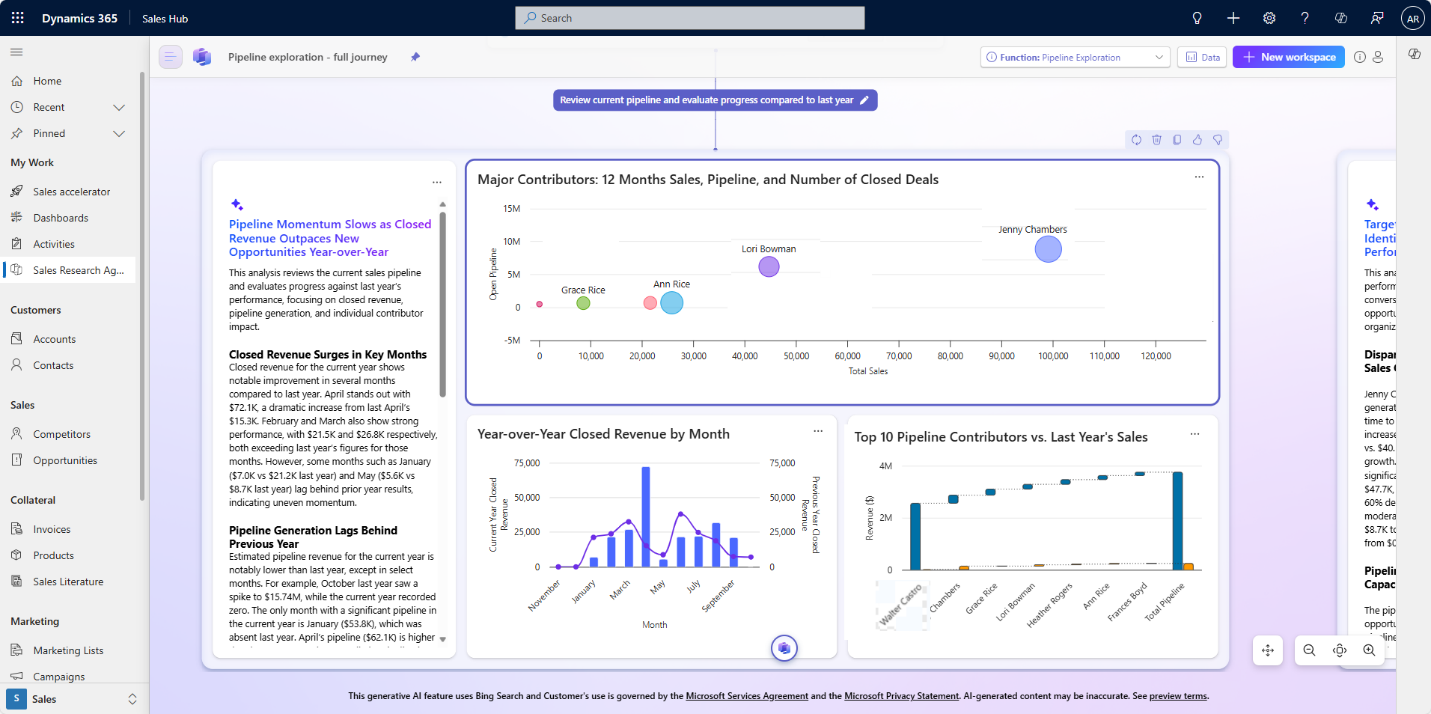}
\caption{The Sales Research Agent in the Dynamics 365 Sales Hub.}
\label{fig:sales_research_agent_dynamics_365}
\end{figure}

This post introduces the architecture and evaluation methodology and results behind Microsoft’s Sales Research Agent. Its technical innovations distinguish the Sales Research Agent from other available offerings, from multi-agent orchestration and multi-model support to advanced techniques for schema intelligence, self-correction and validation. In determining how best to evaluate the Sales Research Agent, Microsoft reviewed existing AI benchmarks and ultimately decided to create the \textbf{Sales Research Bench}, a new benchmark purpose-built to measure the quality of AI-powered Sales Research on business data, in alignment with the business questions, needs, and priorities of sales leaders. In head-to-head evaluations completed on October 19, 2025, the Sales Research Agent outperformed Claude Sonnet 4.5 by 13 points and ChatGPT-5 by 24.1 points on a 100-point scale. 

\begin{figure}[H]
\includegraphics[width=13.76cm,height=9.31cm]{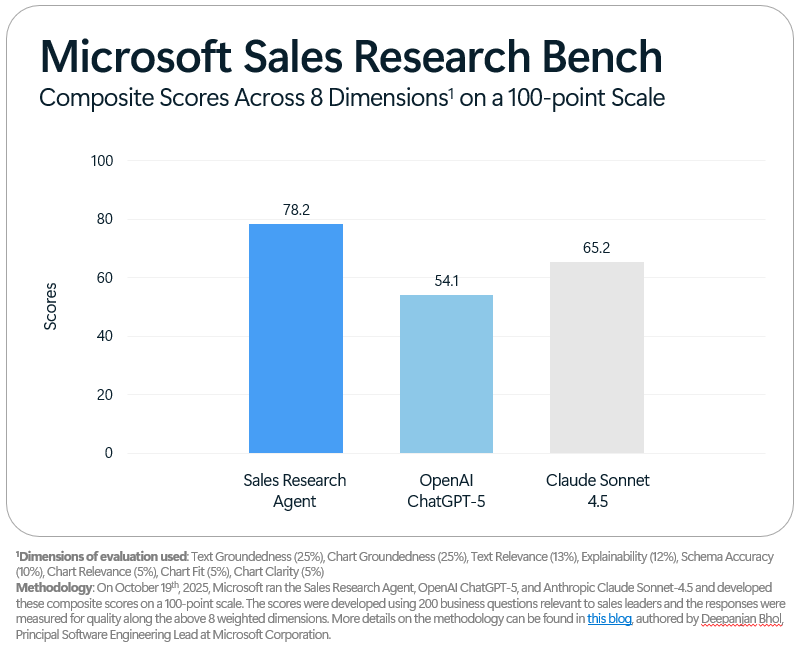}
\caption{Sales Research Bench Composite Score Results.}
\label{fig:sales_research_bench_composite_score}
\end{figure}

\textbf{\textsuperscript{1}Results: }Results reflect testing completed on October 19, 2025, applying the Sales Research Bench methodology to evaluate Microsoft’s Sales Research Agent (part of Dynamics 365 Sales), ChatGPT by OpenAI using a ChatGPT Pro license with GPT-5 in Auto mode, and Claude Sonnet 4.5 by Anthropic using a Claude Max license.

\textbf{Methodology and Evaluation dimensions:} Sales Research Bench includes 200 business research questions relevant to sales leaders that were run on a sample customized data schema. Each AI solution was given access to the sample dataset using different access mechanisms that aligned with their architecture. Each AI solution was judged by LLM judges for the responses the solution generated to each business question, including text and data visualizations. We evaluated quality based on 8 dimensions, weighting each according to qualitative input from customers, what we have heard customers say they value most in AI tools for sales research: Text Groundedness (25$\%$), Chart Groundedness (25$\%$), Text Relevance (13$\%$), Explainability (12$\%$), Schema Accuracy (10$\%$), Chart Relevance (5$\%$), Chart Fit (5$\%$), and Chart Clarity (5$\%$). Each of these dimensions received a score from an LLM judge from 20 as the worst rating to 100 as the best. For example, the LLM judge would give a score of 100 for chart clarity if the chart is crisp and well labeled, score of 20 if the chart is unreadable or misleading. Text Groundedness and Text Relevance used Azure Foundry’s out-of-box LLM evaluators, while judging for the other six dimensions leveraged Open AI’s GPT 4.1 model with specific guidance. A total composite score was calculated as a weighted average from the 8 dimension-specific scores. More details on the methodology can be found in the rest of this blog.

Microsoft will continue to use the evals in Sales Research Bench to drive continuous improvement of the Sales Research Agent, and Microsoft intends to publish the full evaluation package in the coming months, so others can run it to verify published results or benchmark the agents they use (example evals from the benchmark are included in this paper). \textit{}

\section{Sales Research Agent architecture}

The architecture of the Sales Research Agent sets it apart from other offerings, delivering both technical innovation and business value.

\begin{enumerate}
	\item \textbf{Multi-Agent Orchestration:} The Sales Research Agent uses a dynamic multi-agent infrastructure that orchestrates the development of the research blueprints, the text-based narratives and data visualizations accompanied by an explanation of the agent’s work. Specialized agents are invoked at each step in the journey to deliver domain-optimized insights for user questions, taking organizational data as well as business and user context into account.

	\item \textbf{Multi-Model Support:} This multi-agent infrastructure enables each specialized agent to use the model that is best suited to the task at hand. Microsoft tests how each specialized agent performs with different models. Models are easily swapped out to continue optimizing the Sales Research Agent’s quality as the models available evolve over time. 

	\item \textbf{Support for Business Language: }There is a difference between business language (how business users naturally communicate) and natural language (any language that is not code). The Sales Research Agent can give quality answers to prompts in business language, because it breaks down the prompt into multiple sub-questions, building a research plan and using multi-step reasoning over connected data sources. Additionally, the Sales Research Agent is infused with knowledge of the Sales domain, so it can correctly interpret terminology and context that is only implicit to the user’s prompt. 

	\item \textbf{Schema Intelligence:} The Sales Research Agent can handle both out-of-the-box and customized enterprise schemas, adapting to complex, real-world environments. It has sophisticated techniques and heuristics built in to recognize the tables and columns that are relevant to the user query.

	\item \textbf{Self-Correction and Validation:} The Sales Research Agent incorporates advanced auto-correction mechanisms for its generated responses. Whether producing SQL or Python code, the agent leverages sophisticated code correctors capable of iterative refinement—reviewing, validating, and amending outputs as needed. The correction loop begins with a \textbf{fast, non-reasoning model} to identify and fix straightforward issues. If errors persist, the system escalates to a \textbf{reasoning model} and, if required, a \textbf{more powerful model} to ensure deeper contextual understanding and precise correction. This dynamic, multi-model process helps to ensure that the final code is both accurate and reliable, enhancing the overall quality and trustworthiness of the agent’s insights and recommendations.

	\item \textbf{Explainability:} The system tracks every agent interaction and decision, as well as the SQL query and Python code generated to produce the research blueprint. The Sales Research Agent uses this information to help users quickly verify its accuracy and trace its reasoning. Each blueprint includes Show Work, an explanation in simple language for business users, with an advanced view of SQL queries and more details for technical users. 

\end{enumerate}
Figure 3. A high-level diagram of Sales Research Agent’s architecture and how it connects to business workflows. \\ \includegraphics[width=16.51cm,height=12.66cm]{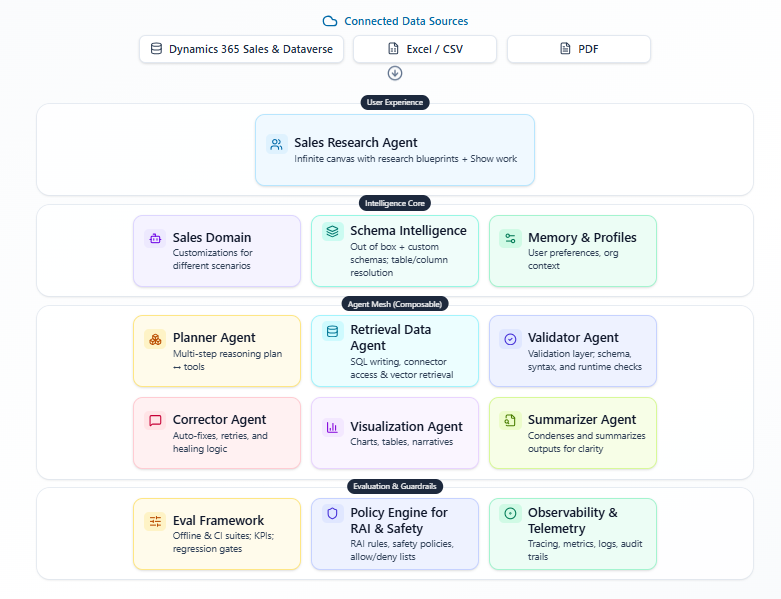}

\section{Why Enterprise Sales Requires a New Evaluation Framework}

In traditional software, unit tests give repeatable proof that core behaviors work and keep working. For AI solutions, evaluations (evals) are needed to demonstrate quality and track continuous improvement over time. 

Enterprises deserve evaluations that are purpose-built for their needs. While there is a wide range of pioneering work on AI evaluation, existing benchmarks miss key attributes that are needed for an AI solution to guide critical business decisions:

\begin{itemize}
	\item The benchmark must reflect the strategic, multi-faceted business questions of sales leaders using their business language.

	\item The benchmark must measure schema accuracy: whether the system correctly handles tables, columns, and joins on system of record schemas that can be highly customized.

	\item The benchmark should assess insights across both text-based narratives and data visualizations, reflecting the outputs with which leaders make decisions.

\end{itemize}
\section{Introducing Sales Research Bench for AI-powered Sales Research}

To meet these demands, Microsoft developed the \textbf{Sales Research Bench,} a composite quality score built to evaluate AI-powered Sales Research solutions in close alignment with customers’ actual questions, environments, and priorities. From engagements with customer sales teams across industries and geographies, Microsoft identified the critical dimensions of quality and created real-world business questions in the language sales leaders use. The data schema on which the evaluations take place is customized to reflect the complexities of customers’ enterprise environments, with their layered business logic and nuanced operational realities. The result is a rigorous benchmark presenting a composite score based on 8 weighted dimensions, as well as dimension-specific scores to reveal where agents excel or need improvement. 

\section{Benchmark Methodology}

The evaluation infrastructure for Sales Research Bench includes:

\begin{itemize}
	\item \textbf{Eval Datasets:} 200 business questions in the language of sales leaders, each associated with its own set of ground-truth answers for validation.

	\item \textbf{Sample enterprise dataset:} Eval questions run on a customized schema, reflecting the complexities of enterprise environments.

	\item \textbf{Evaluators:} LLM-judge-based evaluation, tailored for each of the 8 quality dimensions described below. Azure Foundry out-of-box evaluators are used for Text Groundedness and Text Relevance. For the other 6 dimensions, OpenAI’s GPT 4.1 model is used with specific guidelines on how to score answers, which are provided in the appendix.

\end{itemize}
Here are 3 of the 200 evaluation questions informed by real sales leader questions:

\begin{enumerate}
	\item Looking at closed opportunities, which sellers have the largest gap between Total Actual Sales and Est Value First Year in the ‘Corporate Offices’ Business Segment?

	\item Are our sales efforts concentrated on specific industries or spread evenly across industries?

	\item Compared to my headcount on paper (30), how many people are actually in seat and generating pipeline? 

\end{enumerate}
\subsection{Dimensions of Quality}

The Sales Research Bench aggregates eight dimensions of quality, weighting them as shown in the parentheses below to reflect what we have heard customers say they value most in AI tools for sales research during their  engagements with Microsoft.

\begin{itemize}
	\item \textbf{Text Groundedness (25$\%$)}: Ensures narratives are accurate, faithful to the sample enterprise dataset, and applying correct business definitions. 

	\item \textbf{Chart Groundedness (25$\%$)}: Validates that charts accurately represent the underlying data from the same enterprise dataset.

	\item \textbf{Text Relevance (13$\%$)}: Measures how relevant the insights in the text-based narrative are to the business question. 

	\item \textbf{Explainability (12$\%$)}: Ensures the AI solution accurately and clearly explains how it arrived at its responses.

	\item \textbf{Schema Accuracy (10$\%$):} Verifies the correct selection of tables and columns by evaluating whether the generated SQL query is consistent with the tables, joins, and columns in the ground-truth answers. (Business applications typically consist of approximately 1,000 tables, many featuring around 200 columns, all of which can be highly customized by customers.) 

	\item \textbf{Chart Relevance (5$\%$)}: Validates whether the data and analysis shown in the chart are relevant to the business question. 

	\item \textbf{Chart Fit (5$\%$)}: Evaluates if the chosen visualization matches the analytical need (e.g., line for trends, bar for comparisons). 

	\item \textbf{Chart Clarity (5$\%$)}: Assesses readability, labeling, accessibility, and chart hygiene.

\end{itemize}
Each of these dimensions received a score from an LLM judge from 20 as the worst rating to 100 as the best. For example, the LLM judge would give a score of 100 for chart clarity if the chart is crisp and well labeled, score of 20 if the chart is unreadable or misleading.

\subsection{Sample Enterprise Dataset}

Evaluation needs representative conditions to be useful. Through customer engagements, Microsoft identified numerous edge cases from highly customized schemas, complex joins and filters, and nuanced business logic (like pipeline coverage and attainment calculations). 

For instance, most customers customize their schemas with custom tables and columns, such as replacing an industry column with an industry table, and linking it to the customer object, or adding market and business segment instead of using an existing segment field. As a result, their environments often contain both the out-of-box tables and columns as well as customized tables and fields, all with similar names. By systematically incorporating these edge cases into the sample custom schema, Sales Research Bench evaluates how agents perform outside of the ``happy path" to assess enterprise readiness.

Figure 4. Example evaluation case (see the Appendix for more examples)  \\ \includegraphics[width=16.51cm,height=17.14cm]{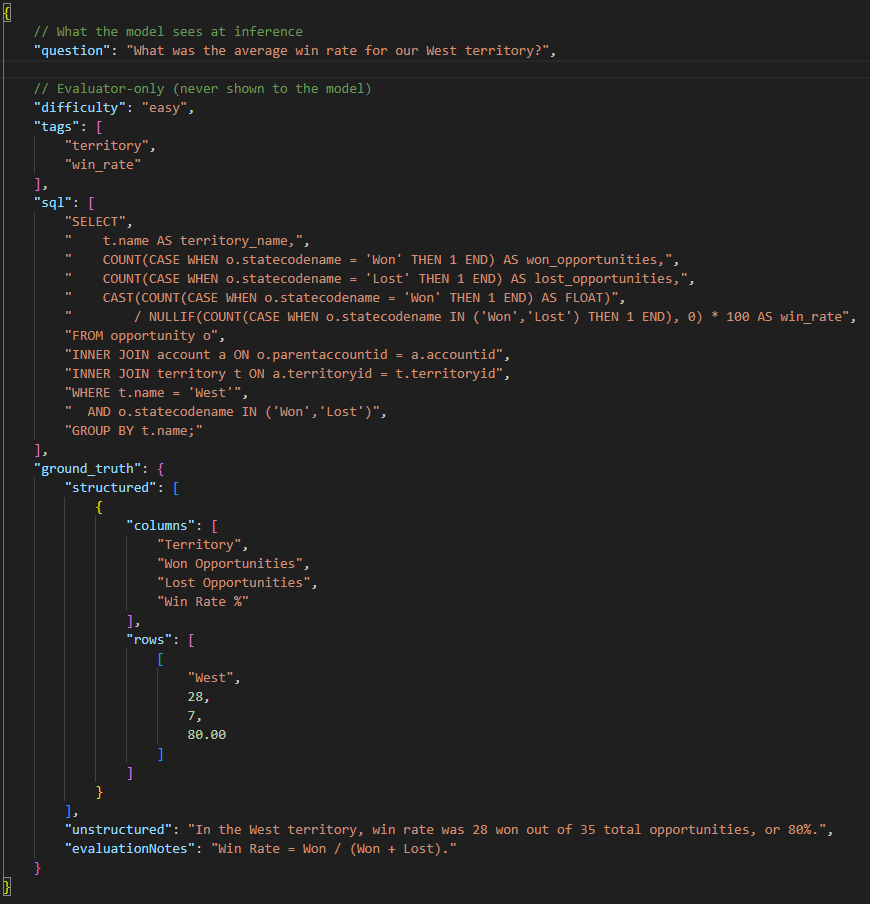}

\subsubsection{Evaluating Sales Research Agent and Other Solutions}

In addition to the Sales Research Agent, Microsoft evaluated ChatGPT by OpenAI using a Pro license with GPT-5 in Auto mode and Claude Sonnet 4.5 by Anthropic using a Max license. The licenses were chosen to optimize for quality: ChatGPT’s pricing page describes Pro as ``full access to the best of ChatGPT," while Claude’s pricing page recommends Max to ``get the most out of Claude."\footnote{ ChatGPT Pricing and Pricing $\vert$ Claude, both accessed on October 19, 2025} Similarly, ChatGPT’s evaluation was run using Auto mode, a setting that allows ChatGPT’s system to determine the best-suited model variant for each prompt. 

Microsoft implemented a controlled evaluation environment where all systems - \textbf{Sales Research Agent}, \textbf{ChatGPT-5}, and \textbf{Claude} \textbf{Sonnet 4.5 }worked with \textbf{identical questions and data}, but through different access mechanisms aligned with their respective architectures. 

The \textbf{Sales Research Agent} has a \textbf{native multi-agent orchestration layer} that connects directly to Dynamics 365 Sales data. This allows it to autonomously discover schema relationships and entity dependencies, and to perform natural-language-to-query reasoning natively within its own orchestration stack.

Since ChatGPT and Claude do not support relational line-of-business source systems out of box, Microsoft enabled access to the \textbf{same dataset} by \textbf{mirroring it into an Azure SQL instance}. Mirroring was done to preserve all the data types, primary keys, foreign keys, and relationships between tables from Dataverse to Azure SQL. This Azure SQL copy was exposed through the \textbf{MCP SQL connector}, ensuring that ChatGPT and Claude retrieved the \textbf{exact same data} but through a standardized external interface. Once responses were captured, they were evaluated using the same evaluators against the exact same evaluation rubrics.

Finally, prompts to ChatGPT and Claude included instructions to create charts and to explain how they got to their answers (Sales Research Agent has this functionality out of box.)
\clearpage  
\section{Evaluation Results}
In a test of 200 evals on the customized schema, Sales Research Agent earned a composite score of 78.2 on a 100-point scale, while Claude Sonnet 4.5 earned 65.2 and ChatGPT-5 earned 54.1.

The chart below presents the Sales Research Bench composite scores, with scores for each dimension overlaid on the bars within the stacked bar chart.

\begin{figure}[H]
\includegraphics[width=10.77cm,height=12.7cm]{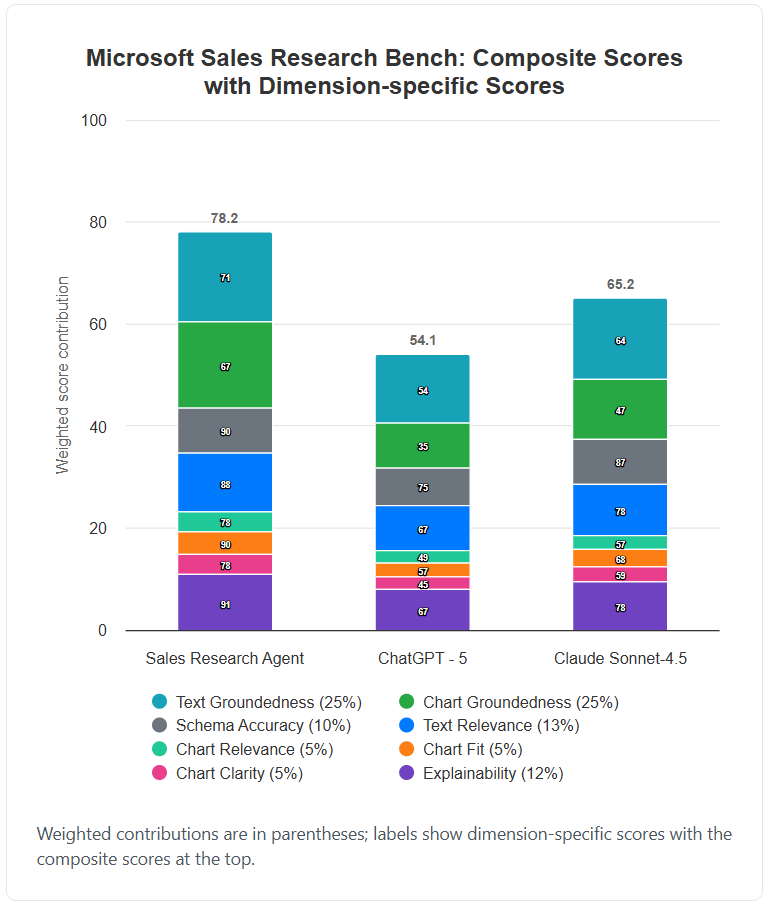}
\caption{Sales Research Bench Composite Scores with Dimension-specific Scores.}
\label{fig:sales_research_bench_composite_scores}
\end{figure}

Breaking this down, the Sales Research Agent outperformed other solutions on all 8 dimensions, with the biggest deltas in chart-related dimensions (groundedness, fit, clarity, and relevance), and the smallest deltas in schema accuracy and text groundedness. Claude Sonnet 4.5 outperformed ChatGPT-5 on all 8 dimensions, with the biggest delta in chart clarity and the smallest delta in chart relevance. 

\begin{figure}[H]
\includegraphics[width=13.76cm,height=6.79cm]{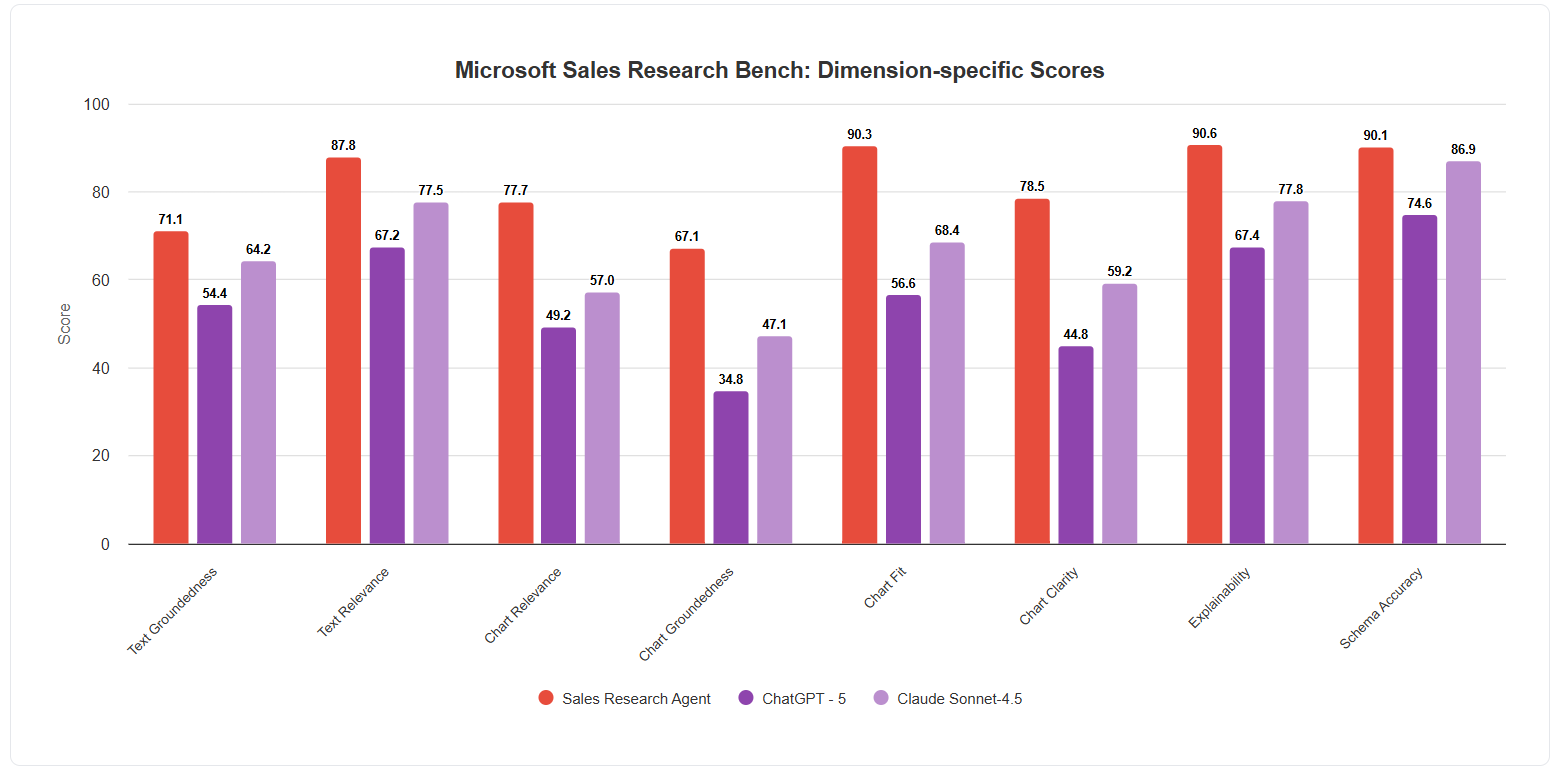}
\caption{Sales Research Bench Scores by Dimension.}
\label{fig:sales_research_bench_scores_dimension}
\end{figure}

\section{Looking Ahead}

Sales Research Agent introduces a new generation AI-first business application that transforms how sales leaders can approach and solve complex business questions. The Sales Research Bench was created in parallel to represent a new standard for enterprise AI evaluation: Rigorous, comprehensive, and aligned with the needs and priorities of sales leaders. 

Upcoming plans for the Sales Research Bench include using the benchmark for continuous improvement of the Sales Research Agent, running further comparisons against a wider range of competitive offerings, and publishing the eval package so customers can run it themselves to verify the published results and benchmark the agents they use. Evaluation is not a one-time event. Scores can be tracked across releases, ensuring that AI solutions evolve to meet customer needs.

Looking beyond Sales Research Bench, Microsoft plans to develop eval frameworks and benchmarks for more business functions and agentic solutions— in customer service, finance, and beyond. The goal is to set a new standard for trust and transparency in enterprise AI.

\section{Appendix:}

\subsection{Scoring Guidelines provided to LLM Judges}

Text Groundedness and Text Relevance used Azure Foundry’s out-of-box LLM evaluators. Below are the guidelines provided to the LLM judges for the other six quality dimensions. These judges leverage Open AI’s GPT 4.1 model.

\subsubsection{Schema accuracy:}

\begin{itemize}
	\item 100: Perfect match - all golden tables and columns are present (extra columns OK, Dynamics equivalents OK)

	\item 80: Very good - minor missing columns or one missing table

	\item 60: Good - some important columns or tables missing but core schema is there

	\item 40: Fair - significant schema differences but some overlap

	\item 20: Poor - major schema mismatch or completely different tables

\end{itemize}
\subsubsection{Explainability:}

\begin{itemize}
	\item 100 (Excellent): Explanation is highly detailed, perfectly describes what the generated SQL does, technically accurate, and provides clear business context

	\item 80 (Good): Explanation is sufficiently detailed and mostly accurate with minor gaps in describing the SQL operations

	\item 60 (Fair): Explanation provides adequate detail but misses some important SQL operations or has minor inaccuracies

	\item 40 (Poor): Explanation lacks sufficient detail to understand the SQL operations or has significant inaccuracies

	\item 20 (Very Poor): Explanation is too vague, mostly incorrect, or provides insufficient detail about the generated SQL

\end{itemize}
\subsubsection{Chart Groundedness:}

\begin{itemize}
	\item 100: Data accurately matches ground truth OR both ground truth $\&$ chart empty

	\item 80: Minor data inaccuracies

	\item 60: Some data inaccuracies

	\item 40: Major data inaccuracies

	\item 20: data completely mismatches ground truth

\end{itemize}
\subsubsection{Chart Relevance:}

\begin{itemize}
	\item 100: Question and chart strongly reinforce each other

	\item 60: Question and chart loosely align but with some disconnect

	\item 20: Question and chart do not align at all

\end{itemize}
\subsubsection{Chart Fit:}

\begin{itemize}
	\item 100: Optimal chart choice for the task OR both ground truth $\&$ chart empty (appropriate emptiness)

	\item 60: Acceptable chart choice but not optimal for the task

	\item 20: inappropriate/confusing chart type

\end{itemize}
\subsubsection{Chart Clarity:}

\begin{itemize}
	\item 100: Chart is crisp and well-labeled

	\item 60: Chart readable but missing labels/clarity elements

	\item 20: Chart unreadable, misleading

\end{itemize}

\subsection{Sample Evaluation Datasets}
Below are some of the evaluation datasets that we have used to benchmark the performance of the Sales Research Agent against all the evaluation rubrics mentioned above. The same questions were also evaluated against competitive offerings. We include three representative items here to show the structure (question, difficulty, SQL, tags, and ground truth).

\begin{verbatim}
{
    "question": "Looking at closed opportunities, which sellers have the largest gap between Total Actual Sales and Est Value First Year in the ‘Corporate Offices’ Business Segment?"",
    "difficulty": "hard",
    "sql": [
        "SELECT su.[fullname] AS [seller_name],",
        "       COUNT(*) AS [closed_deals],",
        "       SUM(CAST(COALESCE(o.[sop_totalactualsales], o.[actualvalue_base]) AS DECIMAL(38,2))) AS [total_actual_sales],",
        "       SUM(CAST(o.[sop_estvaluefirstyear_base] AS DECIMAL(38,2))) AS [total_est_value_first_year],",
        "       SUM(CAST(COALESCE(o.[sop_totalactualsales], o.[actualvalue_base]) AS DECIMAL(38,2)))",
        "         - SUM(CAST(o.[sop_estvaluefirstyear_base] AS DECIMAL(38,2))) AS [sales_gap]",
        "FROM [dbo].[opportunity] AS o",
        "JOIN [dbo].[systemuser] AS su ON CAST(o.[ownerid] AS NVARCHAR(36)) $=$ CAST(su.[systemuserid] AS NVARCHAR(36))",
        "JOIN [dbo].[sop_businesssegment] AS bs ON CAST(o.[sop_businesssegment] AS NVARCHAR(36)) $=$ CAST(bs.[sop_businesssegmentid] AS NVARCHAR(36))",
        "WHERE o.[statecodename] $=$ 'Won' AND bs.[sop_name] $=$ 'Corporate Offices' AND su.[fullname] <> '' AND o.[sop_estvaluefirstyear_base] IS NOT NULL",
        "GROUP BY su.[fullname]",
        "HAVING SUM(CAST(COALESCE(o.[sop_totalactualsales], o.[actualvalue_base]) AS DECIMAL(38,2))) IS NOT NULL",
        "ORDER BY [sales_gap] DESC;"
    ],
    "tags": [
        "seller-performance",
        "variance",
        "actuals-vs-estimate"
    ],
    "ground_truth": {
        "structured": [
            {
                "columns": [
                    "seller_name",
                    "closed_deals",
                    "total_actual_sales",
                    "total_est_value_first_year",
                    "sales_gap"
                ],
                "rows": [
                    [
                        "Jenny Chambers",
                        3,
                        44501.69,
                        16010.15,
                        28491.54
                    ],
                    [
                        "Heather Rogers",
                        1,
                        21501.05,
                        4190.57,
                        17310.48
                    ],
                    [
                        "Grace Rice",
                        1,
                        21223.33,
                        6789.20,
                        14434.13
                    ],
                    [
                        "Ann Rice",
                        1,
                        3243.23,
                        7267.77,
                        -4024.54
                    ]
                ]
            }
        ],
        ``unstructuredtext": "Largest positive gaps: Jenny Chambers (+$28.49K), Heather Rogers (+$17.31K), and Grace Rice (+$14.43K). Ann Rice under-shot estimate (-$4.02K).",
        "evaluationNotes": "Gap $=$ Total Actual Sales $-$ Est First Year; Corporate Offices segment only; closed (Won) opps."
    }
}
{
    "question": "Are our sales efforts concentrated on specific industries or spread evenly across industries?",
    "difficulty": "medium",
    "sql": [
        "SELECT ",
        "    [sop_industry].[sop_name] AS [industry_name],",
        "    COUNT([opportunity].[opportunityid]) AS [total_opportunity_count],",
        "    COUNT(CASE ",
        "              WHEN [opportunity].[statecodename] NOT IN ('Won','Lost','Canceled') ",
        "              THEN 1 ",
        "         END) AS [open_opportunity_count]",
        "FROM ",
        "    [opportunity]",
        "INNER JOIN ",
        "    [account] ON CAST([opportunity].[parentaccountid] AS NVARCHAR(36)) $=$ CAST([account].[accountid] AS NVARCHAR(36))",
        "INNER JOIN ",
        "    [sop_industry] ON CAST([account].[sop_industry] AS NVARCHAR(36)) $=$ CAST([sop_industry].[sop_industryid] AS NVARCHAR(36))",
        "GROUP BY ",
        "    [sop_industry].[sop_name]",
        "HAVING ",
        "    COUNT([opportunity].[opportunityid]) > 0",
        "ORDER BY ",
        "    [open_opportunity_count] DESC;"
    ],
    "tags": [
        "industry",
        "concentration",
        "open-vs-total"
    ],
    "ground_truth": {
        "structured": [
            {
                "columns": [
                    "industry_name",
                    "total_opportunity_count",
                    "open_opportunity_count"
                ],
                "rows": [
                    [
                        "Legal Services",
                        1352,
                        240
                    ],
                    [
                        "Insurance",
                        1210,
                        212
                    ],
                    [
                        "Non-Durable Merchandise Retail",
                        946,
                        177
                    ],
                    [
                        "Inbound Repair and Services",
                        695,
                        126
                    ],
                    [
                        "Outbound Consumer Service",
                        740,
                        124
                    ],
                    [
                        "Design, Direction and Creative Management",
                        719,
                        119
                    ],
                    [
                        "Building Supply Retail",
                        633,
                        118
                    ],
                    [
                        "Durable Manufacturing",
                        569,
                        111
                    ],
                    [
                        "Business Services",
                        597,
                        108
                    ],
                    [
                        "Broadcasting Printing and Publishing",
                        597,
                        104
                    ],
                    [
                        "Accounting",
                        551,
                        104
                    ],
                    [
                        "Distributors, Dispatchers and Processors",
                        562,
                        104
                    ],
                    [
                        "Financial",
                        606,
                        102
                    ],
                    [
                        "Consulting",
                        532,
                        100
                    ],
                    [
                        "Agriculture and Non-petrol Natural Resource Extraction",
                        586,
                        95
                    ],
                    [
                        "Doctor's Offices and Clinics",
                        497,
                        90
                    ],
                    [
                        "Brokers",
                        579,
                        90
                    ],
                    [
                        "Food and Tobacco Processing",
                        489,
                        86
                    ],
                    [
                        "Consumer Services",
                        451,
                        81
                    ],
                    [
                        "Eating and Drinking Places",
                        448,
                        76
                    ],
                    [
                        "Equipment Rental and Leasing",
                        425,
                        74
                    ],
                    [
                        "Entertainment Retail",
                        429,
                        73
                    ],
                    [
                        "Inbound Capital Intensive Processing",
                        419,
                        71
                    ]
                ]
            }
        ],
        ``unstructuredtext": "Effort is broad but skewed: Legal Services and Insurance have the most total opps, while several industries maintain 70–120 open opps.",
        "evaluationNotes": "Counts total vs open opps per industry; ordered by open count.
    }
}
{
    "question": "Compared to my headcount on paper (30), how many people are actually in seat and generating pipeline?",
    "difficulty": "medium",
    "sql": [
        "WITH open_opps AS (",
        "  SELECT o.*",
        "  FROM opportunity o",
        "  WHERE o.statecodename NOT IN ('Won','Lost','Canceled')",
        ")",
        "SELECT",
        "  CAST(30 AS INT) AS headcount_on_paper,",
        "  COUNT(DISTINCT open_opps.ownerid) AS active_pipeline_users,",
        "  (30 - COUNT(DISTINCT open_opps.ownerid)) AS delta_needed,",
        "  (SELECT COUNT(*) FROM opportunity) AS total_opportunities,",
        "  (SELECT COUNT(*) FROM open_opps) AS open_opportunities,",
        "  (SELECT SUM(CAST(o2.estimatedvalue_base AS DECIMAL(38,2))) FROM open_opps o2) AS open_pipeline_value;"
    ],
    "tags": [
        "capacity",
        "headcount",
        "pipeline"
    ],
    "ground_truth": {
        "structured": [
            {
                "columns": [
                    "headcount_on_paper",
                    "active_pipeline_users",
                    "delta_needed",
                    "total_opportunities",
                    "open_opportunities",
                    "open_pipeline_value"
                ],
                "rows": [
                    [
                        30,
                        7,
                        23,
                        14860,
                        2662,
                        16047760.29
                    ]
                ]
            }
        ],
        ``unstructuredtext": "Only 7 sellers have active pipeline against a plan of 30 (shortfall of 23). Open pipeline totals $16.05M across 2,662 opps.",
        "evaluationNotes": "Active sellers counted as distinct owners on current pipeline."
    }
}
\end{verbatim}
\nocite{*}
\bibliographystyle{unsrtnat}  % or plainnat / plain, etc.
\bibliography{references}     % name of your .bib file
\end{document}